\documentclass[runningheads]{llncs}
\usepackage{graphicx}
\usepackage{amsmath,amssymb} 
\usepackage{color}
\usepackage{hyperref}

\usepackage{tabularx,tabulary}
\usepackage{bbm}
\usepackage{subfigure}
\usepackage{comment}
\usepackage{booktabs}
\usepackage{float}
\begin{document}
\title{Graph Distillation for Action Detection with Privileged Modalities} 
\titlerunning{Graph Distillation}
%
\author{Zelun Luo\inst{1,2}\thanks{\scriptsize Work done during an internship at Google Cloud AI.} \and
Jun-Ting Hsieh\inst{1} \and \\
Lu Jiang\inst{2} \and
Juan Carlos Niebles\inst{1,2} \and
Li Fei-Fei\inst{1,2}}
%
\authorrunning{Z. Luo et al.}
%
\institute{Stanford University \and Google Inc.}
\maketitle              

\begin{figure}[ht]
\begin{center}
\includegraphics[width=\linewidth]{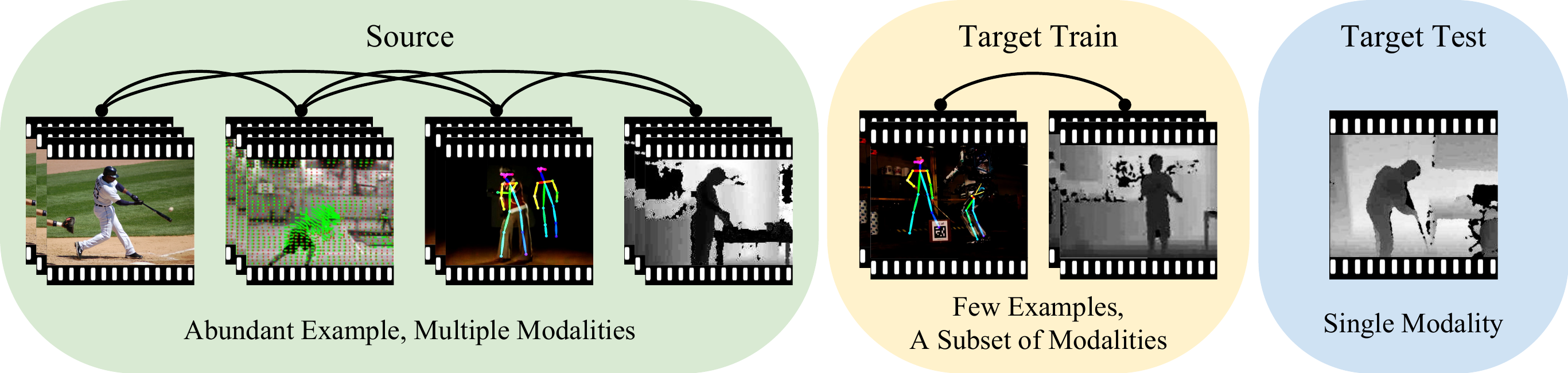}
\end{center}
\caption{\textbf{Our problem statement.} In the source domain, we have abundant data from multiple modalities. In the target domain, we have limited data and a subset of the modalities during training, and only one modality during testing. The curved connectors between modalities represent our proposed graph distillation.}
\label{fig:pull}
\end{figure}

\begin{abstract}
We propose a technique that tackles action detection in multimodal videos under a realistic and challenging condition in which only limited training data and partially observed modalities are available. 
Common methods in transfer learning do not take advantage of the extra modalities potentially available in the source domain. On the other hand, previous work on multimodal learning only focuses on a single domain or task and does not handle the modality discrepancy between training and testing.
In this work, we propose a method termed graph distillation that incorporates rich privileged information from a large-scale multimodal dataset in the source domain, and improves the learning in the target domain where training data and modalities are scarce. 
We evaluate our approach on action classification and detection tasks in multimodal videos, and show that our model outperforms the state-of-the-art by a large margin on the NTU RGB+D and PKU-MMD benchmarks. The code is released at \url{http://alan.vision/eccv18_graph/}.

\end{abstract}

\section{Introduction}
Recent advancements in deep convolutional neural networks (CNN) have been successful in various vision tasks such as image recognition \cite{imagenet,resnet,alexnet} and object detection \cite{fast_rcnn,yolo,faster_rcnn}. A notable bottleneck for deep learning, when applied to multimodal videos, is the lack of massive, clean, and task-specific annotations, as collecting annotations for videos is much more time-consuming and expensive. Furthermore, restrictions such as privacy or runtime may limit the access to only a subset of the video modalities during test time.

The scarcity of training data and modalities is encountered in many real-world applications including self-driving cars, surveillance, and health care. A representative example is activity understanding on health care data that contain Personally Identifiable Information (PII)~\cite{hand_hygiene,senior_home}. On the one hand, the number of labeled videos is usually limited because either important events such as falls~\cite{fall_detection_principles,fall_detection_survey} are extremely rare or the annotation process requires a high level of medical expertise. On the other hand, RGB violates individual privacy and optical flow requires non-real-time computations, both of which are known to be important for activity understanding but are often unavailable at test time. Therefore, detection can only be performed on real-time and privacy-preserving modalities such as depth or thermal videos.

Inspired by these problems, we study action detection in the setting of limited training data and partially observed modalities. To do so, we make use of a large action classification dataset that contains various \emph{heterogeneous} modalities as the source domain to assist the training of the action detection model in the target domain, as illustrated in Fig.~\ref{fig:pull}. Following the standard assumption in transfer learning~\cite{yosinski2014transferable}, we assume that the source and target domain are similar to each other. We define a modality as a privileged modality if (1) it is available in the source domain but not in the target domain; (2) it is available during training but not during testing. 

We identify two technical challenges in this problem. First of all, due to modality discrepancy in types and quantities, traditional domain adaption or transfer learning methods~\cite{subspace_alignment,transfer} cannot be directly applied. Recent work on knowledge and cross-modal distillation~\cite{distillation_hinton,li2017learning,unifying,privileged_on_depth_shi} provides a promising way of transferring knowledge between two models. Given two models, we can specify the distillation as the direction from the strong model to the weak model. With some adaptations, these methods can be used to distill knowledge between modalities. However, these adapted methods fail to address the second challenge: how to leverage the privileged modalities effectively. More specifically, given multiple privileged modalities, the distillation directions and weights are difficult to be pre-specified. Instead, the model should learn to dynamically adjust the distillation based on different actions or examples. 
For instance, some actions are easier to detect by optical flow whereas others are easier by skeleton features, and therefore the model should adjust its training accordingly. However, this dynamic distillation paradigm has not yet been explored by existing methods.

To this end, we propose the novel \emph{graph distillation} method to learn a dynamic distillation across multiple modalities for action detection in multimodal videos. The graph distillation is designed as a layer attachable to the original model and is end-to-end learnable with the rest of the network. The graph can dynamically learn the example-specific distillation to better utilize the complementary information in multimodal data. As illustrated in Fig.~\ref{fig:pull}, by effectively leveraging the privileged modalities from both the source domain and the training stage of the target domain, graph distillation significantly improves the test-time performance on a single modality. Note that graph distillation can be applied to both single-domain (from training to testing) and cross-domain (from one task to another) tasks. For our cross-domain experiment (from action classification to detection), we utilized the most basic transfer learning approach, \textit{i.e.} pre-train and fine-tune, as this is orthogonal to our contributions. We can potentially achieve even better results with advanced transfer learning and domain adaptation techniques and we leave it for future study. 

We validate our method on two public multimodal video benchmarks: PKU-MMD~\cite{pku_mmd} and NTU RGB+D~\cite{ntu_rgbd}. The datasets represent one of the largest public multimodal video benchmarks for action detection and classification. The experimental results show that our method outperforms the state-of-the-art approaches. Notably, it improves the state-of-the-art by 9.0\% on PKU-MMD~\cite{pku_mmd} (at 0.5 tIoU threshold) and by 6.6\% on NTU RGB+D~\cite{ntu_rgbd}. The remarkable improvement on the two benchmarks is a convincing validation of our method. 

To summarize, our contribution is threefold. 
(1) We study a realistic and challenging condition for multimodal action detection with limited training data and modalities. To the best of our knowledge, we are first to effectively transfer multimodal privileged information across domains for action detection and classification.
(2) We propose the novel graph distillation layer that can dynamically learn to distill knowledge across multiple privileged modalities and can be attached to existing models and learned in an end-to-end manner.
(3) Our method outperforms the state-of-the-art by a large margin on two popular benchmarks, including action classification task on the challenging NTU RGB+D~\cite{ntu_rgbd} and action detection task on PKU-MMD~\cite{pku_mmd}.

\section{Related Work}
\noindent\textbf{Multimodal Action Classification and Detection.}
The field of action classification~\cite{i3d_carreira,two_stream_simonyan,c3d_tran} and action detection~\cite{sst_buch_cvpr17,daps,thumos2015,structured_segment_network} in RGB videos has been studied by the computer vision community for decades. The success in RGB videos has given rise to a series of studies on action recognition in multimodal videos~\cite{hbrnn,jiang2014easy,koppula2013learning,li2018visual,cad,wang2012mining}. Specifically, with the availability of depth sensors and joint tracking algorithms, extensive research has been done on action classification and detection in RGB-D videos~\cite{ni2013rgbd,shahroudy2017deep,shao2017performance,yu2016structure} as well as skeleton sequences~\cite{10-stream,lstm_trust_gate,attention_lstm,skeleton_visualization,ntu_rgbd,geometric_features}. Different from previous work, our model focuses on leveraging privileged modalities on a source dataset with abundant training examples. We show that it benefits action detection when the target training dataset is small in size, and when only one modality is available at test time.

\noindent\textbf{Video Understanding Under Limited Data.}
Our work is largely motivated by real-world situations where data and modalities are limited. For example, surveillance systems for fall detection~\cite{fall_detection_principles,fall_detection_survey} often face the challenge that annotated videos of fall incidents are hard to obtain, and more importantly, yhr recording of RGB videos is prohibited due to privacy concerns. Existing approaches to tackling this challenge include using transfer learning~\cite{luo2017label,transfer_learning_survey} and leveraging noisy data from web queries~\cite{chen2015webly,liang2016learning,yeung2017learning}. Specifically to our problem, it is common to transfer models trained on action classification to action detection.

The transfer learning methods are proved to be effective. However, it requires the source and target domains to have the same modalities. In reality, the source domain often contains richer modalities. For instance, suppose the depth video is the only available modality in the target domain, it remains nontrivial to transfer the other modalities (\textit{e.g.} RGB, optical flow) even though they are readily available in the source domain and could make the model more accurate. Our method provides a practical approach to leveraging the rich multimodal information in the source domain, benefiting the target domain of limited modalities.

\noindent\textbf{Learning Using Privileged Information.} Vapnik and Vashist~\cite{privileged_vapnik} introduced a \textit{Student-Teacher} analogy: in real-world human learning, the role of a teacher is crucial to the student's learning process since the teacher can provide explanations, comments, comparisons, metaphors, etc. They proposed a new learning paradigm called Learning Using Privileged Information (LUPI), where at training time, additional information about the training example is provided to the learning model. At test time, the privileged information is not available, and the student operates without the supervision of the teacher~\cite{privileged_vapnik}.

Several work employed privileged information (PI) on SVM classifiers~\cite{privileged_vapnik,hidden_information_wang}. Ding et al.~\cite{ding2015missing} handled missing modality transfer learning using latent low-rank constraint. Recently, the use of privileged information has been combined with deep learning in various settings such as PI reconstruction~\cite{privileged_on_depth_shi,pedestrian_xu}, information bottleneck~\cite{information_bottleneck_motiian}, and Multi-Instance Multi-Label (MIML) learning~\cite{yang2017miml}. The idea more related to our work is the combination of distillation and privileged information, which will be discussed next.

\noindent\textbf{Knowledge Distillation.}
Hinton et al.~\cite{distillation_hinton} introduced the idea of knowledge distillation, where knowledge from a large model is distilled to a small model, improving the performance of the small model at test time. This is done by adding a loss function that matches the outputs of the small network to the high-temperature soft outputs of the large network~\cite{distillation_hinton}. Lopez-Paz et al.~\cite{unifying} later proposed a generalized distillation that combined distillation and privileged information. This approach was adopted by~\cite{hallucination_hoffman} and~\cite{distillation_gupta} in cross-modality knowledge transfer. Our graph distillation method is different from prior work~\cite{distillation_hinton,li2017learning,unifying,privileged_on_depth_shi} in that the privileged information contains multiple modalities and that the distillation directions and weights are dynamically learned rather than being predefined by human experts.

\section{Method}
Our goal is to assist the training in the target domain with limited labeled data and modalities by leveraging the source domain dataset with abundant examples and multiple modalities. We address the problem by distilling the knowledge from the privileged modalities. Formally, we model action classification and detection as an $L$-way classification problem, where a ``background class'' is added in action detection. 

Let $\mathcal{D}_{t} = \{(x_i, y_i)\}_{i=1}^{|\mathcal{D}_{t}|}$ denote the training set in the target domain, where $x_i\in\mathbb{R}^d$ is the input and $y_i\in\mathbb{R}$ is an integer denoting the class label. Since training data in the target domain is limited, we are interested in transferring knowledge from a source dataset $\mathcal{D}_{s} = \{(x_i, \mathcal{S}_i, y_i)\}_{i=1}^{|\mathcal{D}_{s}|}$, where $|\mathcal{D}_{s}| \gg |\mathcal{D}_{t}|$, and the source and target data may have different classes. The new element $\mathcal{S}_i = \{x_i^{(1)},...,x_i^{(|\mathcal{S}|)}\}$ is a set of privileged information about the $i$-th sample, where the superscript indexes the modality in $\mathcal{S}_i$. As an example, $x_i$ could be the depth image of the $i$-th frame in a video and $x_i^{(1)},x_i^{(2)},x_i^{(3)} \in \mathcal{S}_i$ might be RGB, optical flow and skeleton features about the same frame, respectively. For action classification, we employ the standard softmax cross entropy loss:
{\small
\begin{equation}
\label{eq:softmax_xentropy}
\ell_c(f(x_i), y_i) = -\sum_{j=1}^L \mathbbm{1}(y_i =j) \log \sigma(f(x_i)),
\end{equation}
}where $\mathbbm{1}$ is the indicator function and $\sigma$ is the softmax function. The class prediction function $f:\mathbb{R}^d \to [1,L]$ computes the probability for each action class.

In the rest of this section, Section~\ref{sec:previledged_knowledge_distilation} discusses the overall objective of privileged knowledge distillation. Section~\ref{sec:collective} details the proposed graph distillation over multiple modalities.

\subsection{Knowledge Distillation with Privileged Modalities}\label{sec:previledged_knowledge_distilation}

\begin{figure}[t]
\subfigure{\label{fig:modela}}
\subfigure{\label{fig:modelb}}
\subfigure{\label{fig:modelc}}
\begin{center}
\includegraphics[width=\linewidth]{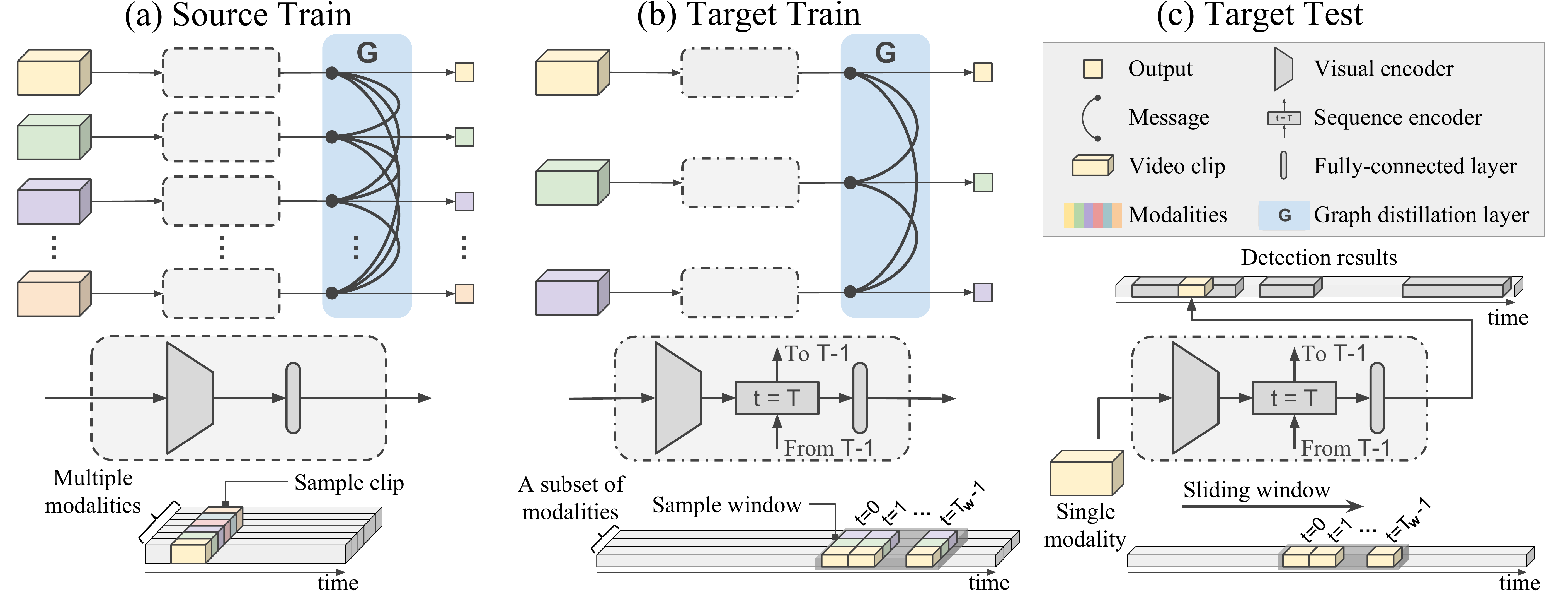}
\end{center}
\caption{\textbf{An overview of our network architectures.} (a) Action classification with graph distillation (attached as a layer) in the source domain. The visual encoders for each modality are trained. (b) Action detection with graph distillation in the target domain at training time. In our setting, the target training modalities is a subset of the source modalities (one or more). Note that the visual encoder trained in the source is transferred and finetuned in the target. (c) Action detection in the target domain at test time, with a single modality.}
\label{fig:model}
\end{figure}

To leverage the privileged information in the source domain data, we follow the standard transfer learning paradigm. We first train a model with graph distillation using all modalities in the source domain, and then transfer only the visual encoders (detailed in Sec~\ref{sec:network_architecture}) of the target domain modalities. Finally, the visual encoder is finetuned with the rest of the target model on the target task. The visual feature encoding step is shared between the tasks in the source and target data and is therefore intuitive to use the same visual encoder architecture (as shown in Fig.~\ref{fig:model}) for both tasks.

To train a graph distillation model on the source data, we minimize:
{\small
\begin{equation}
\label{eq:distilation_loss}
\min \frac{1}{|\mathcal{D}_{s}|} \sum_{(x_i, y_i) \in \mathcal{D}_{s}} \ell_c(f(x_i),y_i) + \ell_m(x_i, \mathcal{S}_i).
\end{equation}}The loss consists of two parts: the first term is the standard classification loss in Eq.~\eqref{eq:softmax_xentropy} and the latter is the imitation loss~\cite{distillation_hinton}. The imitation loss is often defined as the cross-entropy loss on the \emph{soft logits}~\cite{distillation_hinton}. In existing literatures, the imitation loss is computed using a pre-specified distillation direction. For example, Hinton et al.~\cite{distillation_hinton} computed the soft logits by $\sigma(f_{\mathcal{S}}(x_i)/T)$, where $T$ is the temperature, and $f_{\mathcal{S}}$ is the class prediction function of the cumbersome model. Gupta et al.~\cite{distillation_gupta} employed the ``soft logits'' obtained from different layers of the labeled modality. In both cases, the distillation is pre-specified, \textit{i.e.}, from a cumbersome model to a small model in~\cite{distillation_hinton} or from a labeled modality to an unlabeled modality in~\cite{distillation_gupta}. In our problem, 
the privileged information comes from multiple heterogeneous modalities and it is difficult to pre-specify the distillation directions and weights. To this end, our the imitation loss in Eq.~\eqref{eq:distilation_loss} is derived from a dynamic distillation graph.

\subsection{Graph Distillation}\label{sec:collective}

First, consider a special case of graph distillation where only two modalities are involved. 
We employ an imitation loss that combines the logits and feature representation. For notation convenience, we denote $x_i$ as $x_i^{(0)}$ and fold it into $\mathcal{S}_i = \{x_i^{(0)}, \cdots, x_i^{(|\mathcal{S}|)}\}$. Given two modalities $a,b \in [0, |\mathcal{S}|]$ $(a \ne b)$, we use the network architectures discussed in Section~\ref{sec:models} to obtain the logits and the output of the last convolution layer as the visual feature representation.

The proposed imitation loss between two modalities consists of the loss on the logits $l_{logits}$ and the representation $l_{rep}$. The cosine distance is used on both logits and representations as we found the angle of the prediction to be more indicative and better than KL divergence or L1 distance for our problem.

The imitation loss $\ell_m$ from modality $b$ to $a$ is computed by the weighted sum of the logits loss and the representation loss. We encapsulate the loss between two modalities into a message $m_{a \leftarrow b}$ passing from $b$ to $a$, calculated from:
{\small
\begin{equation}
m_{a \leftarrow b}(x_i) = \ell_m(x_i^{(a)}, x_i^{(b)}) = \lambda_1 l_{logits}+\lambda_2 l_{rep},
\label{eq:message_ab}
\end{equation}}where $\lambda_1$ and $\lambda_2$ are hyperparameters. Note that the message is directional, and $m_{a \leftarrow b}(x_i) \ne m_{b \leftarrow a}(x_i)$.

For multiple modalities, we introduce a directed graph of $|\mathcal{S}|$ vertices, named \emph{distillation graph}, where each vertex $v_k$ represents a modality and an edge $e_{k \leftarrow j} \ge 0$ is a real number indicating the strength of the connection from $v_j$ to $v_k$. For a fixed graph, the total imitation loss for the modality $k$ is:
{\small
\begin{equation}
\ell_m(x_i^{(k)}, \mathcal{S}_i) = \sum_{v_j \in \mathcal{N}(v_k)} e_{k \leftarrow j} \cdot m_{k \leftarrow j}(x_i),
\end{equation}}where $\mathcal{N}(v_k)$ is the set of vertices pointing to $v_k$. 

To exploit the dynamic interactions between modalities, we propose to learn the distillation graph along with the original network in an end-to-end manner. Denote the graph by an adjacency matrix $\mathbf{G}$ where $\mathbf{G}_{jk} = e_{k \leftarrow j}$. Let $\phi_k^l$ be the logits and $\phi_k^{l-1}$ be the representation for modality $k$, where $l$ indicates the number of layers in the network. Given an example $x_i$, the graph is learned by:
{\small
\begin{align}
z_i^{(k)}(x_i) &= W_{11} \phi_k^{l-1}(x_i^{(k)}) + W_{12} \phi_k^{l}(x_i^{(k)}), \\
\mathbf{G}_{jk}(x_i) &= e_{k \leftarrow j} = W_{21} [z_i^{(j)}(x_i) \|  z_i^{(k)}(x_i)]
\label{eq:graph_learning}
\end{align}}where $W_{11}$,  $W_{12}$ and $W_{21}$ are parameters to learn and $\cdot \| \cdot$ indicates the vector concatenation. $W_{21}$ maps a pair of inputs to an entry in $\mathbf{G}$. The entire graph is learned by repetitively applying Eq.~\eqref{eq:graph_learning} over all pairs of modalities in $\mathcal{S}$.

As a distillation graph is expected to be sparse, we normalize $\mathbf{G}$ such that the nonzero weights are dispersed over a small number of vertices. Let $\mathbf{G}_{j:} \in \mathbb{R}^{1 \times |\mathcal{S}|}$ be the vector of its $j$-th row. The graph is normalized:
{\small
\begin{equation}
\label{eq:graph_learning_softmax}
\mathbf{G}_{j:}(x_i) = \sigma(\alpha [\mathbf{G}_{j1}(x_i), ..., \mathbf{G}_{j|\mathcal{S}|}(x_i)]),
\end{equation}}where $\alpha$ is used to scale the input to the softmax operator. 

The message passing on distillation graph can be conveniently implemented by attaching a new layer to the original network. As shown in Fig.~\ref{fig:modela}, each vertex represents a modality and the messages are propagated on the graph layer. In the forward pass, we learn a $\mathbf{G} \in \mathbb{R}^{|\mathcal{S}| \times |\mathcal{S}|}$ by Eq.~\eqref{eq:graph_learning} and~\eqref{eq:graph_learning_softmax} and compute the message matrix $\mathbf{M} \in \mathbb{R}^{|\mathcal{S}| \times |\mathcal{S}|}$ by Eq.~\eqref{eq:message_ab} such that $\mathbf{M}_{jk}(x_i)=m_{k \leftarrow j}(x_i)$. The imitation loss to all modalities is calculated by:
{\small
\begin{equation}
\label{eq:message_graph}
\ell_m = (\mathbf{G}(x_i) \odot \mathbf{M}(x_i))^T \mathbf{1},
\end{equation}}where $\mathbf{1} \in \mathbb{R}^{|\mathcal{S}| \times 1}$ is a column vector of ones; $\odot$ is the element-wise product between two matrices; $\mathbf{\ell_m} \in \mathbb{R}^{|\mathcal{S}| \times 1}$ contains imitation loss for every modality in $\mathcal{S}$. In the backward propagation, the imitation loss $\ell_m$ is incorporated in Eq.~\eqref{eq:distilation_loss} to compute the gradient of the total training loss. This graph distillation layer is end-to-end trained with the rest of the network. As shown, the distillation graph is an important and essential structure which not only provides a base for learning dynamic message passing through modalities but also models the distillation as a few matrix operations which can be conveniently implemented as a new layer in the network.

For a modality, its performance on the cross-validation set often turns out to be a reasonable estimator to its contribution in distillation. Therefore, we add a constant bias term $\mathbf{c}$ in Eq.~\eqref{eq:graph_learning_softmax}, where $\mathbf{c} \in \mathbb{R}^{|\mathcal{S}| \times 1}$ and $c_j$ is set w.r.t. the cross-validation performance of the modality $j$ and $\sum_{k=1}^{|\mathcal{S}|} c_k = 1$. Therefore, Eq.~\eqref{eq:message_graph} can be rewritten as:
{\small
\begin{align}
\label{eq:message_graph_final}
\ell_m 
&= ((\mathbf{G}(x_i)+ \mathbf{1} \mathbf{c}^T)\odot\mathbf{M}(x_i))^T \mathbf{1} \\
&= (\mathbf{G}(x_i)\odot\mathbf{M}(x_i))^T\mathbf{1}+(\mathbf{G}_{prior}\odot\mathbf{M}(x_i))^T\mathbf{1}
\end{align}}where $\mathbf{G}_{prior} = \mathbf{1} \mathbf{c}^T$ is a constant matrix. Interestingly, by adding a bias term in Eq.~\eqref{eq:graph_learning_softmax}, we decompose the distillation graph into two graphs: a learned example-specific graph $\mathbf{G}$ and a prior modality-specific graph $\mathbf{G}_{prior}$ that is independent to specific examples. The messages are propagated on both graphs and the sum of the message is used to compute the total imitation loss. There exists a physical interpretation of the learning process. Our model learns a graph based on the likelihood of observed examples to exploit complementary information in $\mathcal{S}$. Meanwhile, it imposes a prior to encouraging accurate modalities to provide more contribution. By adding a constant bias, we use a more computationally efficient approach than actually performing message passing on two graphs. 

So far, we have only discussed the distillation on the source domain. In practice, our method may also be applied to the target domain on which privileged modality is available. In this case, we apply the same method to minimize Eq.~\eqref{eq:distilation_loss} on the target training data. As illustrated in Fig.~\ref{fig:modelb}, a graph distillation layer is added during the training of the target model. At the test time, as shown in Fig.~\ref{fig:modelc}, only a single modality is used.

\section{Action Classification and Detection Models}
\label{sec:models}
In this section, we discuss our network architectures as well as the training and testing procedures for action classification and detection. The objective of action classification is to classify a trimmed video into one of the predefined categories. The objective of action detection is to predict the start time, the end time, and the class of an action in an untrimmed video.

\subsection{Network Architecture}\label{sec:network_architecture}
For action classification, we encode a short clip of video into a feature vector using the visual encoder. For action detection, we first encode all clips in a window of video (a window consists of multiple clips) into initial feature vectors using the visual encoder, then feed these initial feature vectors into a sequence encoder to generate the final feature vectors. For either task, each feature vector is fed into a task-specific linear layer and a softmax layer to get the probability distribution across classes for each clip. Note that a background class is added for action detection. Our action classification and detection models are inspired by~\cite{two_stream_simonyan} and~\cite{montes2016temporal}, respectively. We design two types of visual encoders depending on the input modalities.

\noindent\textbf{Visual Encoder for Images.} Let $X=\{x_t\}_{t=1}^{T_c}$ denote a video clip of image modalities (\textit{e.g.} RGB, depth, flow), where $x_t\in\mathbb{R}^{H\times W\times C}$, $T_c$ is the number of frames in a clip, and $H\times W\times C$ is the image dimension. Similar to the temporal stream in \cite{two_stream_simonyan}, we stack the frames into a $H\times W\times (T_c\cdot C)$ tensor and encode the video clip with a modified ResNet-18~\cite{resnet} with $T_c\cdot C$ input channels and without the last fully-connected layer. Note that we do not use the Convolutional 3D (C3D) network~\cite{i3d_carreira,c3d_tran} because it is hard to train with limited amount of data~\cite{i3d_carreira}.

\noindent\textbf{Visual Encoder for Vectors.} Let $X=\{x_t\}_{t=1}^{T_c}$ denote a video clip of vector modalities (\textit{e.g.} skeleton), where $x_t\in\mathbb{R}^{D}$ and $D$ is the vector dimension. Similar to \cite{10-stream}, we encode the input with a 3-layer GRU network~\cite{gru} with $T_c$ timesteps. The encoded feature is computed as the average of the outputs of the highest layer across time. The hidden size of the GRU is chosen to be the same as the output dimension of the visual encoder for images.

\noindent\textbf{Sequence Encoder.} Let $X = \{x_t\}_{t=1}^{T_c\cdot T_w}$ denote a window of video with $T_w$ clips, where each clip contains $T_c$ frames. The visual encoder first encodes each clip individually into a single feature vector. These $T_w$ feature vectors are then passed into the sequence encoder, which is a 1-layer GRU network, to obtain the class distributions of these $T_w$ clips. Note that the sequence encoder is only used in action detection.

\subsection{Training and Testing}

Our proposed graph distillation can be applied to both action detection and classification. For action detection, we show that our method can optionally pre-train the action detection model on action classification tasks, and graph distillation can be applied in both pre-training and training stages. Both models are trained to minimize the loss in Eq.~\eqref{eq:distilation_loss} on per-clip classification, and the imitation loss is calculated based on the representations and the logits. 

\noindent\textbf{Action Classification.}
Fig.~\ref{fig:modela} shows how graph distillation is applied in training. During training, we randomly sample a video clip of $T_c$ frames from the video, and the network outputs a single class distribution. During testing, we uniformly sample multiple clips spanning the entire video and average the outputs to obtain the final class distribution.

\noindent\textbf{Action Detection.}
Fig.~\ref{fig:modelb} and Fig.~\ref{fig:modelb} show how graph distillation is applied in training and testing, respectively. As discussed earlier, graph distillation can be applied to both the source domain and the target domain. During training, we randomly sample a window of $T_w$ clips from the video, where each clip is of length $T_c$ and is sampled with step size $s_c$. As the data is imbalanced, we set a class-specific weight based on its inverse frequency in the training set. During testing, we uniformly sample multiple windows spanning the entire video with step size $s_w$, where each window is sampled in the same way as training. The outputs of the model are the class distributions on all clips in all windows (potentially with overlaps depending on $s_w$). These outputs are then post-processed using the method in~\cite{montes2016temporal} to generate the detection results, where the activity threshold $\gamma$ is introduced as a hyperparameter.

\section{Experiments}
In this section, we evaluate our method on two large-scale multimodal video benchmarks. The results show that our method outperforms representative baseline methods and achieves the state-of-the-art performance on both benchmarks.

\subsection{Datasets and Setups}\label{sec:dataset_setups}
We evaluate our method on two large-scale multimodal video benchmarks: NTU RGB+D~\cite{ntu_rgbd} (classification) and PKU-MMD~\cite{pku_mmd} (detection). These datasets are selected for the following reasons. (1) They are (one of the) largest RGB-D video benchmarks in each category. (2) The privileged information transfer is reasonable because the domains of the two datasets are similar. (3) They contain abundant modalities, which are required for graph distillation. 

We use NTU RGB+D as our dataset in the source domain, and PKU-MMD in the target domain. In our experiments, unless stated otherwise, we apply graph distillation whenever applicable. Specifically, the visual encoders of all modalities are jointly trained on NTU RGB+D by graph distillation. On PKU-MMD, after initializing the visual encoder with the pre-trained weights obtained from NTU RGB+D, we also learn all available modalities by graph distillation on the target domain. By default, only a single modality is used at test time.

\noindent\textbf{NTU RGB+D~\cite{ntu_rgbd}.} 
It contains 56,880 videos from 60 action classes. Each video has exactly one action class and comes with four modalities: RGB, depth, 3D joints, and infrared. The training and testing sets have 40,320 and 16,560 videos, respectively. All results are reported with cross-subject evaluation.

\noindent\textbf{PKU-MMD~\cite{pku_mmd}.} 
It contains 1,076 long videos from 51 action classes. Each video contains approximately 20 action instances of various lengths and consists of four modalities: RGB, depth, 3D joints, and infrared. All results are evaluated based on the Average Precision (mAP) at different temporal Intersection over Union (tIoU) thresholds between the predicted and the ground truth intervals.

\noindent\textbf{Modalities.} We use a total of six modalities in our experiments: RGB, depth (D), optical flow (F), and three skeleton features (S) named Joint-Joint Distances (JJD), Joint-Joint Vector (JJV), and Joint-Line Distances (JLD)~\cite{ding2017investigation,10-stream}, respectively. The RGB and depth videos are provided in the datasets. The optical flow is calculated on the RGB videos using the dual TV-L1 method~\cite{zach2007duality}. The three spatial skeleton features are extracted from 3D joints using the method in \cite{ding2017investigation} and \cite{10-stream}. Note that we select a subset of the ten skeleton features in~\cite{ding2017investigation,10-stream} to ensure the simplicity and reproducibility of our method, and our approach can potentially perform better with the complete set of features.

\noindent\textbf{Baselines.}
In addition to comparing with the state-of-the-art, we implement three representative baselines that could be used to leverage multimodal privileged information: \textit{multi-task learning}~\cite{caruana1998multitask}, \textit{knowledge distillation}~\cite{distillation_hinton}, and \textit{cross-modal distillation}~\cite{distillation_gupta}. For the multi-task model, we predict the raw pixels of the other modalities from the representation of a single modality, and use the $L_2$ distance as the multi-task loss. For the distillation methods, the imitation loss is calculated as the high-temperature cross-entropy loss on the soft logits~\cite{distillation_hinton}, and $L_2$ loss on both representations and soft logits in cross-modal distillation~\cite{distillation_gupta}. These distillation methods originally only support two modalities, and therefore we average the pairwise losses to get the final loss.

\begin{table}[t]
\centering
\scriptsize
\caption{Comparison with state-of-the-art on NTU RGB+D. Our models are trained on all modalities and tested on the single modality specified in the table. The available modalities are RGB, depth (D), optical flow (F), and skeleton (S).}
\label{ntu_state_of_the_art}
\begin{tabular}{lc@{\hskip 0.1in}c@{\hskip 0.8in}l@{\hskip 0.4in}c@{\hskip 0.1in}c}
\toprule
Method & Test Modality & mAP & Method & Test Modality & mAP  \\
\midrule
Shahroudy~\cite{shahroudy2017deep} & RGB+D & 0.749 & Ours & RGB & \textbf{0.895} \\
Liu~\cite{liu2017viewpoint} & RGB+D & 0.775 & Ours  & D & 0.875 \\
Liu~\cite{skeleton_visualization} & S & 0.800 & Ours  & F & 0.857 \\
Ding~\cite{ding2017investigation} & S & 0.823 & Ours  & S & 0.837 \\
Li~\cite{10-stream} & S & 0.829 &&& \\
\bottomrule
\end{tabular}
\end{table}

\begin{table}[t]
\centering
\scriptsize
\caption{Comparison of action detection methods on PKU-MMD with state-of-the-art models. Our models are trained with graph distillation using all privileged modalities
and tested on the modalities specified in the table. ``Transfer'' refers to pre-training on NTU RGB+D on action classification. The available modalities are RGB, depth (D), optical flow (F), and skeleton (S).}
\label{pku_state_of_the_art}
\begin{tabular}{l@{\hskip 0.1in}c@{\hskip 0.1in}c@{\hskip 0.1in}c@{\hskip 0.1in}c}
\toprule
\multicolumn{2}{c}{} & \multicolumn{3}{c}{mAP @ tIoU thresholds ($\theta$)} \\
\cmidrule(r){3-5}
Method & Test Modality & 0.1 & 0.3 & 0.5 \\ 
\midrule
Deep RGB (DR) \cite{pku_mmd} & RGB & 0.507 & 0.323 & 0.147 \\
Qin and Shelton \cite{pku_result_qin} & RGB & 0.650 & 0.510 & 0.294 \\
Deep Optical Flow (DOF) \cite{pku_mmd} & F & 0.626 & 0.402 & 0.168 \\
Raw Skeleton (RS) \cite{pku_mmd} & S & 0.479 & 0.325 & 0.130 \\
Convolution Skeleton (CS) \cite{pku_mmd} & S & 0.493 & 0.318 & 0.121 \\
Wang and Wang \cite{pku_result_wang_workshop} & S & 0.842 & - & 0.743 \\
RS+DR+DOF \cite{pku_mmd} & RGB+F+S & 0.647 & 0.476 & 0.199 \\
CS+DR+DOF \cite{pku_mmd} & RGB+F+S & 0.649 & 0.471 & 0.199 \\
\midrule
Ours (w/o $|$ w/ transfer) & RGB & 0.824 $|$ 0.880 & 0.813 $|$ 0.868 & 0.743 $|$ 0.801 \\
Ours (w/o $|$ w/ transfer) & D   & 0.823 $|$ 0.872 & 0.817 $|$ 0.860 & 0.752 $|$ 0.792 \\
Ours (w/o $|$ w/ transfer) & F   & 0.790 $|$ 0.826 & 0.783 $|$ 0.814 & 0.708 $|$ 0.747 \\
Ours (w/o $|$ w/ transfer) & S   & 0.836 $|$ 0.857 & 0.823 $|$ 0.846 & 0.764 $|$ 0.784 \\
Ours (w/ transfer) & RGB+D+F+S & \bf{0.903} & \bf{0.895} & \bf{0.833} \\
\bottomrule
\end{tabular}
\end{table}

\noindent\textbf{Implementation Details.} 
For action classification, we train the visual encoder from scratch for 200 epochs using SGD with momentum with learning rate $10^{-2}$ and decay to $10^{-1}$ at epoch 125 and 175. $\lambda_1$ and $\lambda_2$ are set to $10,5$ respectively in Eq.~\eqref{eq:message_ab}. At test time we sample 5 clips for inference. For action detection, the visual and sequence encoder are trained for 400 epochs. The visual encoder is trained using SGD with momentum with learning rate $10^{-3}$, and the sequence encoder is trained with the Adam optimizer~\cite{kingma2015adam} with learning rate $10^{-3}$. The activity threshold $\gamma$ is set to $0.4$. For both tasks, we down-sample the frame rates of the datasets by a factor of 3. The clip length and detection window $T_c$ and $T_w$ are both set to 10. For the graph distillation, $\alpha$ is set to 10 in Eq.~\eqref{eq:graph_learning_softmax}. The output dimensions of the visual and sequence encoder are both set to 512. Since it is nontrivial to jointly train on multiple modalities from scratch, we employ curriculum learning~\cite{bengio2009curriculum} to train the distillation graph. To do so, we first fix the distillation graph as an identity matrix (uniform graph) in the first 200 epochs. In the second stage, we compute the constant vector $\mathbf{c}$ in Eq.~\eqref{eq:message_graph_final} according to the cross-validation results, and then learn the graph in an end-to-end manner.

\subsection{Comparison with State-of-the-Art}\label{sec:exp_soa}

\begin{figure}[t]
\begin{center}
\includegraphics[width=\linewidth]{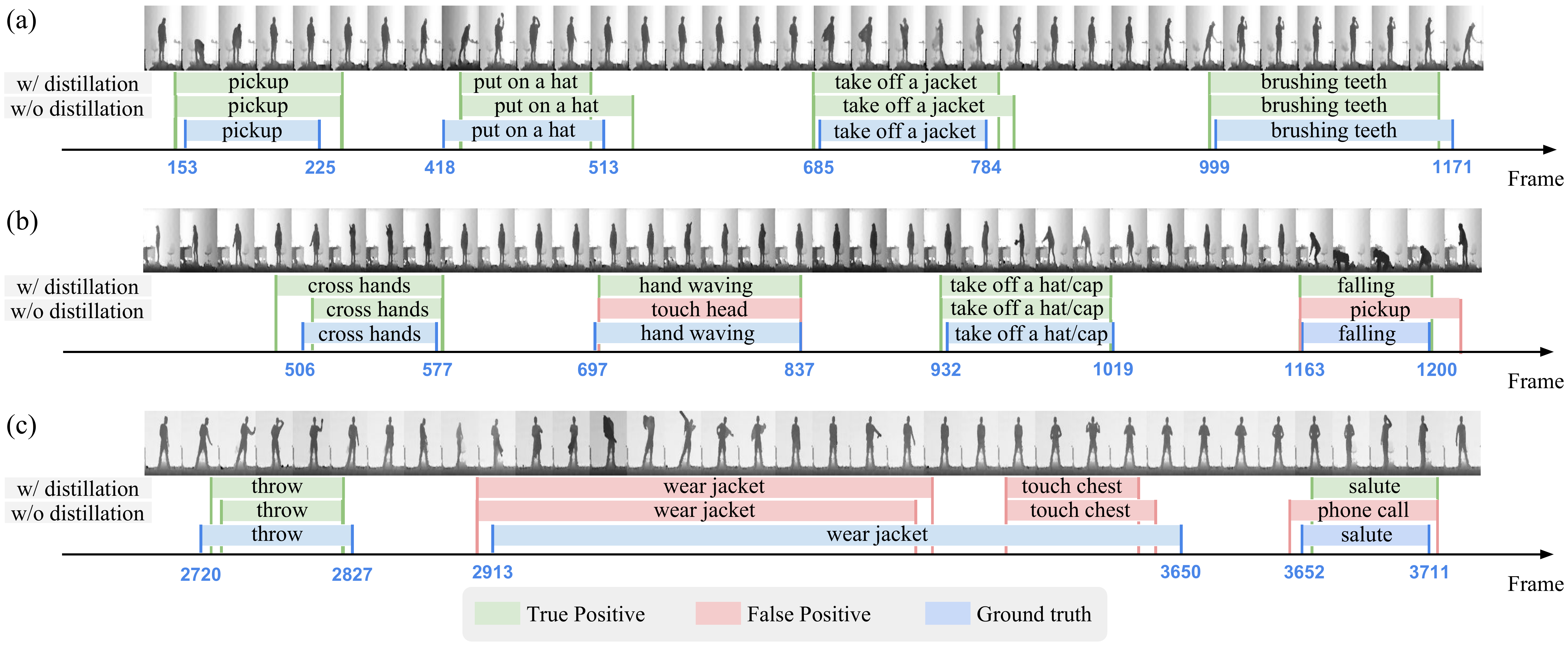}
\end{center}
\caption{\textbf{A comparison of the prediction results on PKU-MMD.} (a) Both models make correct predictions. (b) The model without distillation in the source makes errors. Our model learns motion and skeleton information from the privileged modalities in the source domain, which helps the prediction for classes such as ``hand waving'' and ``falling''. (c) Both models make reasonable errors.}
\label{fig:detection}
\end{figure}

\begin{figure}[t]
\begin{center}
\includegraphics[width=0.8\linewidth]{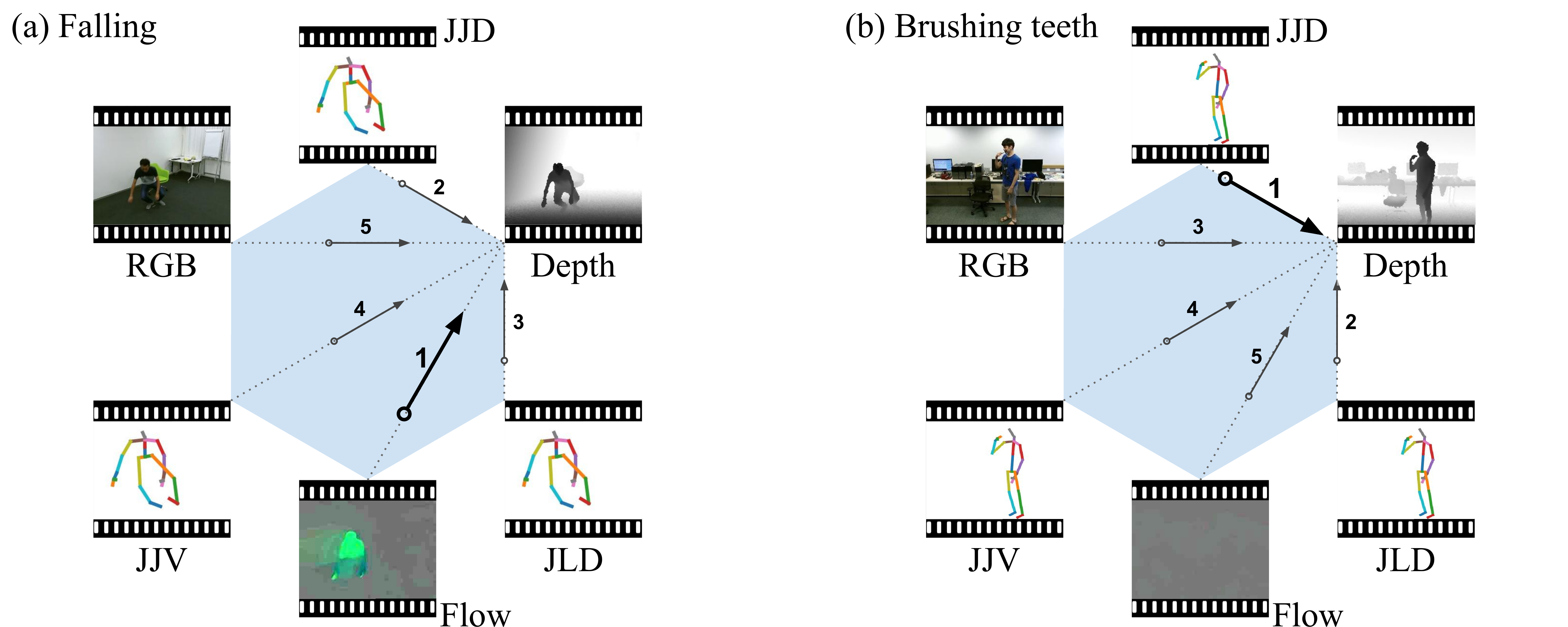}
\end{center}
\caption{\textbf{The visualization of graph distillation on NTU RGB+D.} The numbers indicate the ranks of the distillation weights, with 1 being the largest and 5 being the smallest. (a) Class ``falling'': Our graph assigns more weight to optical flow because optical flow captures the motion information. (b) Class ``brushing teeth'': In this case, motion is negligible, and our graph assigns the smallest weight to it. Instead, it assigns the largest weight to skeleton data.}
\label{fig:graph}
\end{figure}

\noindent\textbf{Action Classification.} Table~\ref{ntu_state_of_the_art} shows the comparison of action classification with state-of-the-art models on NTU RGB+D dataset. Our graph distillation models are trained and tested on the same dataset in the source domain. NTU RGB+D is a very challenging dataset and has been recently studied in numerous studies~\cite{10-stream,liu2017viewpoint,skeleton_visualization,luo2017unsupervised,shahroudy2017deep}. Nevertheless, as we see, our model achieves the state-of-the-art results on NTU RGB+D. It yields a 4.5\% improvement, over the previous best result, using the depth video and a remarkable 6.6\% using the RGB video. After inspecting the results, we found the improvement mainly attributes to the learned graph capturing complementary information across multiple modalities. Fig.~\ref{fig:graph} shows example distillation graphs learned on NTU RGB+D. The results show that our method, without transfer learning, is effective for action classification in the source domain.

\noindent\textbf{Action Detection.} Table~\ref{pku_state_of_the_art} compares our method on PKU-MMD with previous work. Our model outperforms existing methods across all modalities. The results substantiate that our method can effectively leverage the privileged knowledge from multiple modalities. Fig.~\ref{fig:detection} illustrates detection results on the depth modality with and without the proposed distillation.

\subsection{Ablation Studies on Limited Training Data}\label{sec:ablation}
Section~\ref{sec:exp_soa} has shown that our method achieves the state-of-the-art results on two public benchmarks. However, in practice, the training data are often limited in size. To systematically evaluate our method on limited training data, as proposed in the introduction, we construct mini-NTU RGB+D and mini-PKU-MMD by randomly sub-sampling 5\% of the training data from their full datasets and use them for training. For evaluation, we test the model on the full test set.

\begin{table}[t]
\centering
\scriptsize
\caption{The comparison with (a) baseline methods using Privileged Information (PIs) on mini-NTU RGB+D, (b) distillation graphs on mini-NTU RGB+D and mini-PKU-MMD. Empty graph trains each modality independently. Uniform graph uses a uniform weight in distillation. Prior graph is built according to the cross-validation accuracy of each modality. Learned graph is learned by our method. ``D'' refers to the depth modality.}
\subtable[\label{ntu_baselines}Baseline methods using PIs.]
{
  \renewcommand{\arraystretch}{1.1}
  \begin{tabular}{lcc}
  \toprule
  Method & mAP / RGB \\
  \midrule
  Empty graph & 0.464 \\
  Multi-task \cite{caruana1998multitask}  & 0.456 \\
  Cross-distillation \cite{distillation_gupta}  & 0.503 \\
  Knowledge distillation \cite{distillation_hinton}  & 0.524 \\
  Learned graph & \bf{0.619} \\
  \bottomrule
  \end{tabular}
}
\subtable[\label{different_graphs}Different distillation graphs.]{
  \begin{tabular}{l@{\hskip 0.1in}c@{\hskip 0.2in}c}
  \toprule
  \multicolumn{1}{c}{} & mini-NTU & mini-PKU \\
  \cmidrule(r){2-3}
  Graph & {\tiny mAP / RGB} & {\tiny mAP @ 0.5 / D} \\
  \midrule
  Empty graph & 0.464 & 0.501 \\
  Uniform graph & 0.537 & 0.513 \\
  Prior graph & 0.571 & 0.515 \\
  Learned graph & \bf{0.619} & \bf{0.559}\\
  \bottomrule
  \end{tabular}
}
\end{table}

\begin{table}[t]
\centering
\scriptsize
\caption{The mAP comparison on mini-PKU-MMD at different tIoU threshold $\theta$. The depth modality is chosen for testing. ``src'', ``trg'', and ``PI'' stand for source, target, and privileged information, respectively.}
\label{pku_mmd_baselines}
\begin{tabular}{c@{\hskip 0.2in}l@{\hskip 0.4in}c@{\hskip 0.4in}c@{\hskip 0.4in}c}
\toprule
\multicolumn{2}{c}{} & \multicolumn{3}{c}{mAP @ tIoU thresholds ($\theta$)} \\
\cmidrule(r){3-5}
 & Method & $0.1$ & $0.3$ & $0.5$ \\
\midrule
1&trg only & 0.248 & 0.235 & 0.200 \\
2&src + trg & 0.583 & 0.567 & 0.501 \\
3&src w/ PIs + trg & 0.625 & 0.610 & 0.533 \\
4&src + trg w/ PIs & 0.626 & 0.615 & 0.559 \\
5&src w/ PIs + trg w/ PIs & 0.642 & 0.629 & 0.562 \\
\midrule
6&src w/ PIs + trg & 0.625 & 0.610 & 0.533  \\
7&src w/ PIs + trg w/ 1 PI & 0.632 & 0.615 & 0.549 \\
8&src w/ PIs + trg w/ 2 PIs & 0.636 & 0.624 & 0.557 \\
9&src w/ PIs + trg w/ all PIs & 0.642 & 0.629 & 0.562 \\
\bottomrule
\end{tabular}
\end{table}

\noindent\textbf{Comparison with Baseline Methods.} Table~\ref{ntu_baselines} shows the comparison with the baseline models that uses privileged information (see Section~\ref{sec:dataset_setups}). The fact that our method outperforms the representative baseline methods validates the efficacy of the graph distillation method.

\noindent\textbf{Efficacy of Distillation Graph.} Table \ref{different_graphs} compares the performance of predefined and learned distillation graphs. The proposed learned graph is compared with an empty graph (no distillation), a uniform graph of equal weights, and a prior graph computed using the cross-validation accuracy of each modality. Results show that the learned graph structure with modality-specific prior and example-specific information obtains the best results on both datasets.

\noindent\textbf{Efficacy of Privileged Information.} Table~\ref{pku_mmd_baselines} compares our distillation and transfer under different training settings. The input at test time is a single depth modality. By comparing row 2 and 3 in Table~\ref{pku_mmd_baselines}, we see that when transferring the visual encoder to the target domain, the one pre-trained with privileged information in the source domain performs better than its counterpart. As discussed in Section~\ref{sec:collective}, graph distillation can also be applied to the target domain. By comparing row 3 and 5 (or row 2 and 4) of Table~\ref{pku_mmd_baselines}, we see that performance gain is achieved by applying the graph distillation in the target domain. The results show that our graph distillation can capture useful information from multiple modalities in both the source and target domain.

\noindent\textbf{Efficacy of Having More Modalities.} The last three rows of Table \ref{pku_mmd_baselines} show that performance gain is achieved by increasing the number of modalities used as the privileged information. Note that the test modality is depth, the first privileged modality is RGB, and the second privileged modality is the skeleton feature JJD. The results also suggest that these modalities provide each other complementary information during the graph distillation.

\subsection{Graph Distillation on UCF-101}

In this section, we consider graph edge distillation, a special case of graph distillation on UCF-101~\cite{soomro2012ucf101} in which only two modalities (RGB and optical flow) are available. Table~\ref{ucf101} shows the action classification results on UCF-101 using the two-stream architecture proposed in~\cite{two_stream_simonyan}. The optical flow modality performs significantly better than RGB when training from scratch. This is consistent with previous findings that dense optical flow is able to achieve very good performance in spite of limited training data \cite{two_stream_simonyan}. To testify our method, we train a model on the RGB modality from scratch with distillation. Our distilled model performs much better than the model directly trained from scratch. Note that our distilled model outperforms the fine-tuning model that uses pretrained weights on ImageNet.

\begin{table}[ht]
\scriptsize
\centering
\caption{Action classification results on UCF101. For graph distillation model, we distill knowledge from the optical flow stream to the RGB stream.}
\label{ucf101}
\begin{tabular}{l@{\hskip 0.1in}c@{\hskip 0.2in}c@{\hskip 0.1in}c}
\toprule
Method & Test Modality & mAP & Diff. \\
\midrule
From scratch   & Flow & 0.803 & - \\
From scratch   & RGB & 0.484 & +0.000 \\
ImageNet pretrained & RGB & 0.728 & +0.244 \\
Graph distillation & RGB & \textbf{0.757} & \textbf{+0.273} \\
\bottomrule
\end{tabular}
\end{table}

\section{Conclusion}
This paper tackles the problem of action classification and detection in multimodal video with limited training data and partially observed modalities. We propose the novel graph distillation method to assist the training of the model by leveraging privileged modalities dynamically. Our model outperforms representative baseline methods and achieves the state-of-the-art for action classification on NTU RGB+D dataset and action detection on the PKU-MMD. A direction for future work is to combine graph distillation with advanced transfer learning and domain adaptation techniques.

\section{Acknowledgement} 
This work was supported in part by Stanford Computer Science Department and Clinical Excellence Research Center. We specially thank Li-Jia Li, De-An Huang, Yuliang Zou, and all the anonymous reviewers for their valuable comments.

\clearpage

\bibliographystyle{splncs04}
\bibliography{egbib}

\end{document}